\documentclass{article}

\PassOptionsToPackage{numbers, sort&compress}{natbib}

 \usepackage[dblblindworkshop, final]{neurips_2026}

\makeatletter
\renewcommand{\@noticestring}{}
\makeatother

\usepackage[table,svgnames,x11names]{xcolor}
\definecolor{citecolor}{HTML}{0071bc}
\usepackage[pagebackref=false,breaklinks=true,letterpaper=true,colorlinks,citecolor=citecolor,bookmarks=false]{hyperref}
\usepackage[utf8]{inputenc} 
\usepackage[T1]{fontenc}    
\usepackage{url}            
\usepackage{booktabs}       
\usepackage{amsfonts}       
\usepackage{nicefrac}       
\usepackage{microtype}      

\usepackage{graphicx}       
\usepackage{amsmath}        
\usepackage{multirow}       
\usepackage{tabularx}       
\usepackage{xspace}         
\usepackage{subcaption}     
\usepackage{makecell}       
\usepackage{float}          

\usepackage{enumitem} 

\newcommand{\sysname}{AdaptiveSpec\xspace}

\newcommand{\dcs}{DCS\xspace}
\newcommand{\nsteps}{n_{\text{steps}}}
\newcommand{\topk}{\text{top-}k}
\newcommand{\ndt}{\mathrm{ndt}}

\newcommand{\repourl}{https://anonymous.4open.science/r/adaptive-spec-0B47}

\workshoptitle{LIGHT: Deployable Small Foundation Models}

\title{Margins, Not Windows: Training-Free Per-Step Lossy Speculative Decoding}

\author{Oszkár Urbán\textsuperscript{1} \quad
  Young D. Kwon\textsuperscript{1,2}\thanks{Corresponding author: \texttt{yd.kwon@samsung.com}} \quad
  Stylianos I. Venieris\textsuperscript{2} \quad
  Cecilia Mascolo\textsuperscript{1} \\
  \textsuperscript{1}University of Cambridge \quad
  \textsuperscript{2}Samsung AI Center-Cambridge, UK \\
  }

\begin{document}

\maketitle

\begin{abstract}
Speculative decoding accelerates LLM inference by drafting candidate tokens and verifying them in parallel. Tree-attention drafters such as EAGLE-3 are widely adopted, yet typically hold two decisions fixed: (1) a strict token-match verification rule and (2) a static draft-tree shape. Prior work relaxes each in isolation under limiting assumptions: long draft chains for training-free lossy verification, and adaptive tree shaping under a fixed token budget. We introduce \sysname, a training-free per-step speculative decoding method that adapts both decisions from internal signals already produced during decoding. A per-step margin rule promotes a mismatched draft-proposed token when the ratio of the target's probability on the drafted token to its top-1 probability exceeds a threshold with no dependence on draft length or underlying drafter architecture. A per-step tree policy adjusts the draft tree's depth, width, and node count directly from a fused signal of draft top-1 confidence and a rolling acceptance history capturing recent draft-target agreement, allowing the total draft count to vary rather than only be redistributed. The two adaptations operate on orthogonal axes and compound in effect. Implemented on the SGLang production-grade serving engine, \sysname improves throughput over the state-of-the-art autoregressive speculative decoding method EAGLE-3 by up to 56\%, retaining 94-96\% of downstream task accuracy on average per target model. Evaluated across GSM8K, MATH-500, and HumanEval on three target models (\texttt{DeepSeek-R1-Distill-Llama-8B}, \texttt{Llama-3.1-8B-Instruct}, \texttt{Qwen3-8B}). 



\end{abstract}

\vspace{1.5em}
\section{Introduction}\label{sec:introduction}

Large language models (LLMs) are now widely adopted across natural language processing and beyond, solving diverse tasks~\citep{agarwal-etal-2025-gpt,kamath-etal-2025-gemma}. Their autoregressive design, however, fundamentally limits inference efficiency: each token is produced by a forward pass that must wait for the preceding token. This cost grows as models scale, and is especially pronounced for chain-of-thought reasoning models~\citep{deepseekr1}, which emit long intermediate ``thinking'' traces, often longer than the final answer itself and disregarded once the final answer is produced.

Speculative decoding~\citep{leviathan2023speculative,chen2023accelerating} addresses this autoregressive limitation while preserving the target model's output distribution. It follows a draft-then-verify pattern: a lightweight \emph{draft} model proposes several candidate tokens, and the \emph{target} model verifies them in parallel within a single forward pass, generating correctly predicted tokens at the cost of a single forward pass rather than one pass per token.
Tree-attention drafters~\citep{li2024eagle2,ringel2026ddtree} are a widely adopted implementation of this pattern: at each step, they propose a tree of candidate draft tokens allowing for several continuations.

Recent work improves speculative decoding on two fronts: relaxing the verifier acceptance rule, and adjusting the draft tree shape.
On the \emph{verifier} side, methods loosen the acceptance criterion from strict token match toward semantic equivalence, increasing throughput at moderate accuracy cost: training-required verifier methods~\citep{bachmann2025judge,dong2026beyond} learn an external module from labelled or generated data, while the training-free method, FLy~\citep{li2025fly}, considers a mismatched token semantically equivalent only when a downstream window of the next $W$ tokens (typically $W{=}6$) contains no further mismatch. On EAGLE-3~\citep{li2025eagle3}, however, per-token acceptance collapses with draft depth, to below 10\% by depths 7--9, so a window of this size is very likely to contain a mismatch. As a result, FLy's relaxation on EAGLE-3 yields little net speedup and can fall below the static baseline.
On the \emph{drafter} side, a complementary line adapts the shape and size of the draft tree, given the observation that confident steps benefit more from a deep chain, while uncertain ones need a wide tree of alternatives. TALON~\citep{liu2026talon}, the closest representative, adapts tree shape online from a confidence measure and produces deeper trees on easy steps and wider ones on hard steps. The adaptation, however, happens within a total token budget, and TALON redistributes a node count between depth and width instead of adjusting the draft tree depth and width $(\nsteps, \topk)$ directly.

\vspace{0.5em}
In this work, we address both limitations by introducing \sysname, a training-free, per-step adaptive speculative decoding framework driven entirely by internal signals already computed during decoding.
First, we propose a \textbf{novel per-step lossy verification rule} that promotes a mismatched draft token when the ratio of the target's probability on it to the target's top-1 probability exceeds a threshold, achieving higher throughput at minimal accuracy cost without requiring auxiliary training or an external verifier to judge semantic equivalence.
Second, we propose \textbf{dynamic draft tree-shaping} that directly adjusts the hyperparameters controlling the shape of the draft tree based on the draft's top-1 probability and a rolling acceptance history, without growing under a fixed budget.
In addition, we integrate the two axes into a single solution, \sysname, compounding into further throughput speedup. Our main contributions are summarised as follows:

\begin{enumerate}[leftmargin=2em,itemsep=4pt,topsep=4pt]
\item We propose a novel, training-free, lossy verification rule that promotes a mismatched draft token using only the ratio of the target's probability on it to the target's top-1 probability at the mismatch position, with no requirements on draft chain length, applicable to any drafter.

\item We introduce a dynamic tree-shaping policy that adjusts all three tree-shape drafter hyperparameters directly per decoding step, driven by combining draft confidence with a rolling acceptance rate.

\item We combine these components into \sysname{}, a unified, training-free, internal-signal-driven adaptive speculative decoding method, that improves throughput by 16-45\% on average per target model over the EAGLE-3 baseline, while retaining 94-96\% of downstream task accuracy on average per target model. Both our proposed components of \sysname independently outperform state-of-the-art (SOTA) speculative decoding in their own regime.

\end{enumerate}

\vspace{1em}
\section{Related work}\label{sec:related-work}

\paragraph{Speculative decoding.} Speculative
decoding~\citep{leviathan2023speculative,chen2023accelerating} breaks the
sequential dependency of autoregressive decoding with a draft-and-verify approach. A fast \emph{draft} model proposes $\gamma$ candidate tokens, and the \emph{target} model verifies them in a single forward pass. Because verification runs in parallel, up to $\gamma$ accepted draft tokens are committed per target forward pass. Tree-shaped drafters extend this pattern by proposing a \emph{tree} of candidates per step rather than a chain, exposing multiple continuations to the same verify pass and raising the expected acceptance length \citep{miao2024specinfer,cai2024medusa,ankner2024hydra,du2024glide,xiao2024parallelspec}. We adopt the SOTA autoregressive EAGLE-3~\citep{li2025eagle3} as our drafter. Each EAGLE-3 decoding step alternates two phases: 

\begin{itemize}[leftmargin=2em,itemsep=2pt,topsep=4pt]
    \item \textbf{Draft}: the draft model performs $nsteps$ forwards, branching $topk$ ways per node and pruning to a verify budget of $ndt$ tokens; 
    \newpage
    \item \textbf{Verify}: the target model accepts the proposed tokens consistent with its own distribution. The triplet $(\nsteps, \topk, \ndt)$ thus controls both the draft tree shape and the per-step compute amount.
\end{itemize}

\paragraph{Adaptive draft trees.} A recent body of work has shown that no single draft tree is optimal across a generation; some steps benefit from a deep chain, while others from a wide branching tree. To address this, several works have focused on adapting the tree at runtime \citep{zhang2024adaeagle,xiong2024dyspec,huang2025specdecpp,mamou2024dsl,huo2025c2t,liu2025pearl,gao2025falcon}. EAGLE-2~\citep{li2024eagle2} reranks which tokens fill a fixed-size tree budget. OPT-Tree~\citep{wang2025opttree} re-solves the tree topology at every decoding step to maximise expected acceptance length within a fixed node budget, while Sequoia~\citep{chen2024sequoia} optimises the tree offline via dynamic programming for a given draft--target pair and hardware setup.

\vspace{0.5em}
Closest to our dynamic tree-shaping axis is TALON~\citep{liu2026talon}, which adapts the tree shape online via a confidence-gated expansion rule: after a fixed top-$K$ initialisation at the root (to mitigate early rejection), deeper layers retain and further expand only those candidate tokens whose draft probability lies within a threshold ratio of the layer's most-confident token, growing the tree until a fixed global token budget is reached. This yields topologies that are deep where the draft is confident and wide where it is not. The limitation is that adaptation happens within that global token budget: TALON only redistributes a total node count across depth and width and cannot directly adjust the total draft compute. In addition, as it has not been integrated into a performant inference engine such as SGLang, it remains unclear whether TALON's performance gains would be maintained in a highly optimised production setup.

\vspace{0.5em}
Our \sysname addresses both limitations directly. It jointly adapts all three EAGLE-3 draft-tree hyperparameters $(\nsteps, \topk, \ndt)$ per decoding step from a confidence signal combining the draft model's top-1 probability and a running acceptance rate. Because the candidate triplets span different total node counts, the policy genuinely contracts draft compute on weak-draft steps rather than only redistributing a budget. We implement this inside SGLang, retaining its production-grade performance.

\paragraph{Lossy verification.} Existing work has investigated how to relax the strict acceptance rule to admit mismatches the target considers near-equivalent, raising acceptance rate at a small accuracy cost ~\citep{sun2025blockverification,garipov2025autojudge,wang2025tba}. Judge Decoding~\citep{bachmann2025judge} and \mbox{SemanticSpec}~\citep{dong2026beyond} showed that an external verifier, a judge classifier or a semantic probe over target hidden states, identifies and accepts such mismatches while preserving competitive task accuracy; their main limitation is that this module must be \emph{trained} and therefore comes with limited transferability to new datasets and tasks. 

\vspace{0.5em}
The training-free FLy~\citep{li2025fly} instead promotes a mismatch when an entropy gate condition passes \emph{and} a lookahead window of the next $W$ tokens shows no further mismatch, designed so that the lookahead window can replace a trained verifier.

\vspace{0.5em}
The lookahead window builds on the idea that when LLMs are conditioned on genuinely incorrect tokens, they tend to self-correct in subsequent generation steps ~\citep{pan2023selfcorrection} (a phenomenon referred to as the erroneous token pattern in Judge Decoding \citep{bachmann2025judge}). At a mismatch at position $j$, FLy implements this with an entropy gate followed by a lookahead window of $W$ tokens; if no further mismatches occur within that window, the original token is interpreted as semantically equivalent and retained. Otherwise, a subsequent mismatch signals a self-correction, triggering a rollback.

\vspace{0.5em}
However, this lookahead window does not transfer cleanly to all speculative drafters (\S\ref{sec:windowfree}). In particular, when the mean acceptance rate is low, subsequent mismatches are not necessarily evidence of the target course-correcting an earlier error but rather a consequence of loose draft-target coupling. Additionally, the window is also structurally incompatible with drafters such as EAGLE-3, whose mean acceptance length falls short of the long draft chains ($K{=}15$--$25$) that the FLy window $W$ rule expects.

\vspace{0.5em}
Our proposed \sysname sidesteps both problems by reading the target's probability distribution at the mismatch position: a direct, single-position measurement, independent of accepted token length and of any high-level self-correction idea. It requires no lookahead and no trained module, only the target and draft probability distributions that the verify pass already maintains. Unlike Judge Decoding and SemanticSpec, it incurs no training cost; unlike FLy, it makes no structural assumption about draft length or self-correction pattern.


\section{Method}\label{sec:method}

\paragraph{Problem formulation.} 
A large language model generates one token per forward pass: token $x_t$ is sampled from the conditional distribution $p_\theta(x_t \mid x_{<t})$ over the vocabulary, given preceding tokens $x_{<t} = (x_1, \dots, x_{t-1})$. Generation is therefore sequential; each step must wait for the one before it to finish, therefore decoding latency scales linearly with output length ~\citep{pope2022efficiently}. This sequential dependency is the bottleneck that speculative decoding aims to solve.

\subsection{Overview}
\sysname is a single system with two adaptive axes, one for each phase of EAGLE-3's decoding step (draft and verify), driven only by internal model signals already produced during the decode step and requiring no auxiliary training. The first axis reshapes the draft tree at every step, driven by a draft-confidence signal; the second relaxes the verification rule using the target's probability-distribution margin.
\begin{itemize}[leftmargin=2em,itemsep=4pt,topsep=4pt]
    \item \textbf{Draft phase} (§\ref{sec:method-dynamic}): a Draft Confidence Score (\dcs{}) selects the tree shape $(\nsteps, \topk, \ndt)$ per step, contracting compute on uncertain drafts and expanding it on confident ones.
    \item \textbf{Verify phase} (§\ref{sec:method-lossy}): a margin-based rule promotes drafted tokens at the first mismatch position when the target's distribution indicates the disagreement is semantically close, recovering acceptances that the strict exact-match rule would discard.
\end{itemize}

\begin{figure*}
    \centering
    \includegraphics[width=1\linewidth]{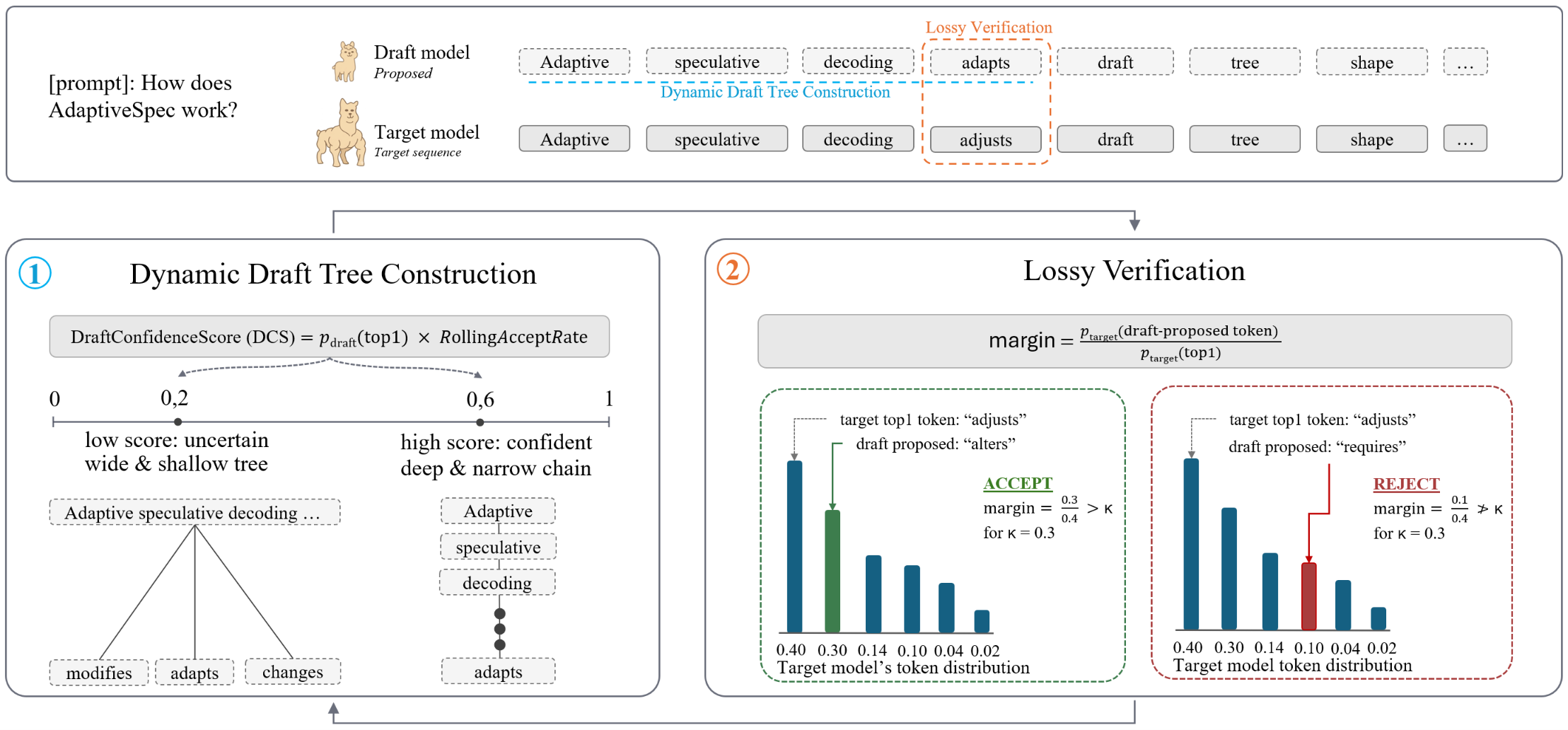}
    \caption{Overview of \sysname. Each decoding step adapts along two axes driven by signals the model already produces. \textbf{(Bottom Left)} During drafting, the Draft Confidence Score (\dcs{}), the product of the draft's top-1 probability and a running acceptance rate (RAR), selects a $(\nsteps, \topk, \ndt)$ triplet: confident steps get a deep narrow chain, uncertain steps a wide shallow tree. \textbf{(Bottom Right)} During verification, at the first mismatch position the target margin rule promotes the drafted token when it exceeds $\kappa$, accepting semantically equivalent disagreements that lossless decoding would discard.}
    \label{fig:placeholder}
\end{figure*}

\subsection{Margin-based lossy verification}
\label{sec:method-lossy}
Lossless speculative decoding in its original form accepts a drafted token only on exact match with the target, discarding even continuations that are semantically equivalent and unnecessarily lowering acceptance rates. To address this, we propose a margin-based rule that enables lossy, yet controlled, verification. We define the margin rule at the first mismatch position $j$ as the ratio of the probability the target model assigned the draft-proposed token to its own top-1 probability choice. Formally:  
\begin{equation}
\mathrm{margin}(j) \;=\;
\frac{p_{\text{target}}(\text{draft}_j)}{p_{\text{target}}(\text{top1}_j)}
\;\in\;[0,1],
\label{eq:margin}
\end{equation}
We define the acceptance rule that promotes the drafted token when $\mathrm{margin}(j) \ge \kappa$, and rejects otherwise. A margin near $1$ means the target nearly chose the drafted token (a semantically equivalent disagreement, safe to promote), while a margin near $0$ means the target was confident that the draft was wrong. The proposed rule is parametrised with threshold $\kappa$ to provide tunable control over the relaxation-strictness during verification.

Our method effectively overcomes the main limitation of FLy's lookahead-window verifier (§\ref{sec:windowfree}); it reads the target's distribution at the mismatched token position, a direct single-position measurement, independent of chain depth and of any high-level self-correction pattern. It requires no lookahead and no trained module, only target probabilities that the verify phase already computes.



\subsection{Dynamic draft tree-shaping}
\label{sec:method-dynamic}
Holding the draft tree shape fixed across decoding is wasteful, since speculative steps are not uniformly hard: confident steps admit a deep chain, while uncertain ones benefit from a wider tree that covers plausible alternatives. EAGLE-3 controls the tree with three hyperparameters, namely, branching factor $\topk$, depth $\nsteps$, and verify-time token budget $\ndt$, all fixed upon deployment.

\sysname selects the values of the triplet via a Draft Confidence Score (\dcs{}), which is computed once per decoding step:
\begin{equation}
\mathrm{DCS}_t \;=\; p_{\text{draft}}(\text{top1}_t)\cdot\mathrm{RAR}_t,
\label{eq:dcs}
\end{equation}

where $p_{\text{draft}}(\text{top1}_t)$ is the draft model's top-1 softmax probability at decode step $t$, and $\mathrm{RAR}_t$ is the rolling acceptance rate (EMA, $\alpha=0.3$) of the fraction of drafted tokens the target accepted, serving as a measure of acceptance history. Before mapping, the score is rescaled by a divisor $d$, so that $\mathrm{DCS}$ saturates to $1$ once the draft confidence $p_{\text{draft}}(\text{top1})\cdot\mathrm{RAR}$ reaches $d$. A smaller $d$ saturates the score sooner and biases the policy toward the deep-narrow configuration. We show a search over $d$ hyperparameter value per model and benchmark in Table~\ref{tab:divisor-sweep}. 

Depth interpolates linearly with \dcs{} and branching with $1{-}\dcs{}$ over a range $(\nsteps, \topk, \ndt) = (3,4,4) \leftrightarrow (7,1,8)$ : a confident draft is given a deep narrow chain, an uncertain draft a wide shallow tree, following the convention adopted by prior work~\citep{liu2026talon}. Crucially, the range spans different total node counts, so the policy genuinely shrinks draft compute on weak-draft steps rather than only redistributing a fixed budget. A circuit breaker reverts to the starting configuration for any request with five consecutive zero-accept steps.

Production-grade inference engines such as SGLang rely on CUDA graphs, which requires static tensor shapes and therefore the values of $\topk$, $\nsteps$, and $\ndt$. To preserve this acceleration under a per-step policy, we pre-capture at startup one CUDA graph for every $(\nsteps, \topk, \ndt)$ triplet the policy could select and swap between them during decoding.

\subsection{\sysname}
\label{sec:method-unified}

\sysname combines the two axes within a single decoding step. During drafting, \dcs{} selects the tree shape $(\nsteps, \topk, \ndt)$, adjusting how the candidate tree is built. During verification, the margin-based rule decides promotion at proposed mismatches. Each axis acts on a different sub-step of the speculative decoding step; tree-shaping during drafting, the margin rule during verification, and their gains compound into increased throughput. Both are driven by signals the decode loop already produces with no extra computation. 


\section{Experiments}\label{sec:experiments}

\subsection{Experimental setup}

\paragraph{Models.}
We evaluate across three target models; \texttt{Llama-3.1-8B-Instruct} \cite{dubey2024llama3}, \texttt{DeepSeek-R1-Distill-Llama-8B}~\citep{deepseekr1}, and \texttt{Qwen3-8B}~\citep{qwen3}, each paired with its publicly released EAGLE-3 draft model. The pairings are shown in Appendix~\ref{sec:appendixpairs}.

\paragraph{Evaluation.} Benchmarks span mathematical
reasoning GSM8K~\citep{cobbe2021gsm8k} and MATH-500~\citep{lightman2024math500}, and code generation HumanEval~\citep{chen2021humaneval} tasks. All main results in Table ~\ref{tab:main} use greedy decoding (temperature $T{=}0$) at batch size~1 on a single NVIDIA A100 (80\,GB), executed on SGLang~\citep{zheng2024sglang} and benchmarked with the SpecForge~\citep{specforge2025} framework. We additionally report a temperature $T{=}1$ analysis and a batch-size sweep in Section~\ref{sec:ablation-hyperparams}, with full table shown in Appendices~\ref{sec:temperature} and~\ref{sec:bsfull}.


\paragraph{Metrics.}
We report output throughput speedup relative to vanilla autoregressive decoding, and mean accepted tokens $\tau$, adding task accuracy recovery for lossy methods, defined as a ratio of task accuracy retained relative to the lossless EAGLE-3. 


\paragraph{Baselines.}
We compare \sysname against three SOTA methods spanning the lossless and lossy regimes: EAGLE-3~\citep{li2025eagle3} for static lossless speculative decoding; an adapted TALON~\citep{liu2026talon} implementation for dynamic draft-tree lossless decoding; and an adapted FLy~\citep{li2025fly} implementation for lossy decoding with a static draft tree. Adapted implementations of TALON and FLy were necessary because the aforementioned works are not currently supported in the SGLang inference engine, and because of drafter limitations, as explained in Section~\ref{sec:windowfree}. The hyperparameters for EAGLE-3 were selected as (3,1,4), matching the configuration reported in the EAGLE-3 SGLang experiments, while dynamic draft tree shapes span the range from $(3,4,4) {\leftrightarrow} (7,1,8)$, following the configurations present in SpecForge's benchmark table~\citep{specforge2025}.

As the reference point for all speedup measurements, we report throughput speedup against standard autoregressive decoding, in which the target model produces a single token per forward pass.

\paragraph{Implementation details.}
We re-implement both FLy and TALON on SGLang using the EAGLE-3 drafters so that all baselines share the same highly optimised inference engine and drafter family.

\textbf{FLy reimplementation on SGLang} FLy's promotion window $w=6$ was designed for $70$B-$405$B target models paired with deep draft chains ($\nsteps = 15$-$25$) (window ratio 24-40\%). We adopt this by decreasing the window size to $w=3$ given the generally shorter mean accepted tokens length as described in Section ~\ref{sec:windowfree}, maintaining a window-to-draft-depth ratio of $\approx 40\%$. Matching this setup by lengthening the draft chain would not be possible, as FLy's authors themselves highlight, pushing the draft length to $15$-$25$ tokens on lightweight drafters degrades draft quality and causes the target to reject drafted tokens substantially more often. 


\textbf{TALON reimplementation on SGLang} TALON's original formulation gates per draft-tree layer, growing the tree node-by-node under a global token budget $\ndt$. We preserve TALON's defining property, budget-constrained adaptive shape selection driven by a confidence-margin threshold $\mu$, but apply it at our per-decode-step granularity. At each step we apply TALON's rule to identify the retained set $\mathcal{P}$, the candidates within the hyperparameter $\mu$ value. When $|\mathcal{P}|$ is large, the policy selects shallow and wide, when small, it selects deep and narrow across the $(\nsteps, \topk)$ configuration. We use the paper's default $\mu = 0.03$. The detailed experiment configuration for each run can be found in Appendix~\ref{sec:experimentconfigs}.


\subsection{Main results}

\begin{table}[t]
\centering
\caption[Main results across the three target models and benchmarks evaluating \sysname{}]{Main results across the three target models and benchmarks evaluating \sysname{}. \sysname{} (Combined) attains the best speedup in every (model, benchmark) cell, improving over the EAGLE-3 baseline by $12$-$56\%$. All \sysname{} lossy results use a single global margin threshold $\kappa{=}0.2$. Speedup is reported relative to vanilla autoregressive decoding; MAT\,$\tau$ is the mean accepted token length; Rec.\,\% is the task accuracy retained relative to lossless EAGLE-3 $(3,1,4)$. Best per-cell speedup and $\tau$ in the \sysname{} model block is in bold. $^{*}$TALON and FLy are evaluated using our SGLang re-implementations on the EAGLE-3 drafter.}
\label{tab:main}
\footnotesize
\setlength{\tabcolsep}{2pt}
\begin{tabular*}{\linewidth}{@{\extracolsep{\fill}}ll ccc ccc ccc@{}}
\toprule
& & \multicolumn{3}{c}{\textbf{GSM8K}} & \multicolumn{3}{c}{\textbf{MATH-500}} & \multicolumn{3}{c}{\textbf{HumanEval}} \\
\cmidrule(lr){3-5}\cmidrule(lr){6-8}\cmidrule(lr){9-11}
\textbf{Model} & \textbf{Method} & Speedup & MAT\,$\tau$ & Rec.\% & Speedup & MAT\,$\tau$ & Rec.\% & Speedup & MAT\,$\tau$ & Rec.\% \\
\midrule
\multirow{6}{*}{Llama-3.1-8B}
 & EAGLE-3                 & 1.93 & 2.83 & - & 1.38 & 2.08 & - & 1.81 & 2.55 & - \\
 \cmidrule(l){2-11}
  & TALON*                   & 2.11 & 3.73 & - & 1.90 & 3.19 & -  & 2.30 & 3.77 & - \\
 & \sysname (Dyn-only) & 2.22 & 3.80 & - & 2.05 & 3.28 & - & 2.43 & 3.99 & - \\
\cmidrule(l){2-11}
 & FLy*                     & 2.17 & 4.06 & 95  & 1.83 & 3.26 & 88  & 2.60 & 4.63 & 92  \\
 & \sysname (Lossy-only) & 2.28 & \textbf{4.35} & 93  & 1.85 & 3.33 & 86  & 2.75 & \textbf{4.89} & 94  \\
 \cmidrule(l){2-11}
 & \sysname (Combined)     & \textbf{2.43} & 4.32 & 94 & \textbf{2.15} & \textbf{3.51} & 90 & \textbf{2.82} & 4.80 & 103 \\
\midrule
\multirow{6}{*}{Qwen3-8B}
 & EAGLE-3                 & 1.68 & 2.39 & - & 1.56 & 2.21 & - & 1.50 & 2.10 & - \\
 \cmidrule(l){2-11}
  & TALON*                   & 1.77 & 2.89 & - & 1.49 & 2.41 & -  & 1.53 & 2.46 & - \\
 & \sysname (Dyn-only) & 1.91 & 3.03 & - & 1.64 & 2.54 & - & 1.72 & 2.65 & - \\
  \cmidrule(l){2-11}
 & FLy*                     & 1.60 & 2.83 & 99  & 1.26 & 2.23 & 82  & 1.34 & 2.37 & 93  \\
 & \sysname (Lossy-only) & 1.61 & 2.85 & 98  & 1.34 & 2.38 & 82  & 1.49 & 2.64 & 93  \\
 \cmidrule(l){2-11}
 & \sysname (Combined)     & \textbf{1.95} & \textbf{3.17} & 99 & \textbf{1.74} & \textbf{2.72} & 82 & \textbf{1.82} & \textbf{2.90} & 104 \\
\midrule
\multirow{6}{*}{DeepSeek-R1-8B}
 & EAGLE-3                 & 2.28 & 3.27 & - & 2.15 & 3.02 & - & 2.05 & 2.86 & - \\
 \cmidrule(l){2-11}
  & TALON*                   & 2.88 & 4.98 & -  & 2.52 & 4.23 & - & 2.26 & 3.77 & -  \\
 & \sysname (Dyn-only) & 2.90 & 4.98 & - & 2.64 & 4.32 & - & 2.41 & 3.90 & - \\
  \cmidrule(l){2-11}
 & FLy*                     & 3.07 & 5.59 & 99  & 2.57 & 4.56 & 92  & 2.32 & 4.12 & 82  \\
 & \sysname (Lossy-only) & \textbf{3.20} & \textbf{5.78} & 96  & 1.80 & 3.28 & 111 & 2.58 & 4.55 & 82  \\
 \cmidrule(l){2-11}
  & \sysname (Combined)     & \textbf{3.20} & 5.66 & 97 & \textbf{2.97} & \textbf{5.01} & 98 & \textbf{2.74} & \textbf{4.60} & 88 \\

\bottomrule
\end{tabular*}

\end{table}



Table~\ref{tab:main} reports throughput speedup, mean accepted tokens length $\tau$, and task-accuracy recovery for lossy methods, on the three target models across GSM8K, MATH-500, and HumanEval.


\paragraph{Dynamic draft tree adaptation improves throughput at full
accuracy retention.}
\sysname{} (Dynamic-only) reshapes the draft tree per step from internal signals and improves over static EAGLE-3 on every (model, benchmark) cell at a lossless setting. The largest single-cell gain is on Llama-3.1-8B MATH-500, where Dynamic-only reaches $2.05\times$ against the static $1.38\times$ ($+49\%$). On average across the three benchmarks, Dynamic-only improves over EAGLE-3 by $+30\%$ on Llama-3.1-8B ($2.23\times$ vs.\ $1.71\times$), $+11\%$ on Qwen3-8B ($1.76\times$ vs.\ $1.58\times$), and $+23\%$ on DeepSeek-R1-8B ($2.65\times$ vs.\ $2.16\times$).

\paragraph{Margin-rule lossy verification outperforms FLy while maintaining accuracy recovery.}

\sysname{} (Lossy-only) exceeds FLy on throughput in all 9 cells except DeepSeek-R1-8B MATH-500 while maintaining task accuracy recovery. The clearest single-cell advantages are on HumanEval, where Lossy-only attains $1.49\times$ vs.\ FLy's $1.34\times$ on Qwen3-8B and $2.58\times$ vs.\ $2.32\times$ on DeepSeek-R1-8B, both $+11\%$. On average, Lossy-only exceeds FLy on Llama-3.1-8B ($2.29\times$ vs.\ $2.20\times$, $+4\%$) and Qwen3-8B ($1.48\times$ vs.\ $1.40\times$, $+6\%$). 


\paragraph{Combined leverages both axes and outperforms all baselines.}
\sysname (Combined) achieves the best speedup on every (model, benchmark) cell, improving over the static EAGLE-3 baseline by $12$-$56\%$ across the cells. The largest single-cell improvement is on Llama-3.1-8B HumanEval, where Combined reaches $2.82\times$ against EAGLE-3's $1.81\times$ ($+56\%$) at $103\%$ accuracy retention. On average, Combined exceeds EAGLE-3 by $+44\%$ on Llama-3.1-8B ($2.47\times$ vs.\ $1.71\times$), $+16\%$ on Qwen3-8B ($1.84\times$ vs.\ $1.58\times$), and $+38\%$ on DeepSeek-R1-8B ($2.97\times$ vs.\ $2.16\times$), the highest average speedup. 




\subsection{Analysis and Ablation study}
\label{sec:ablation-hyperparams}

\paragraph{Ablation study.} We show the contribution of each adaptation axis with a $2{\times}2$ ablation crossing tree
shaping (static, dynamic) with verification (strict, lossy), averaged across GSM8K, MATH-500,
and HumanEval in Table~\ref{tab:2x2}.

On average, each axis improves over the EAGLE-3 baseline (Static$\times$Strict) in isolation.
Dynamic tree shaping lifts the average speedup by $+21.4\%$ ($1.82\times \to 2.21\times$) at no
accuracy cost.
Static lossy verification reaches a $+15.4\%$ speedup ($1.82\times \to 2.10\times$) at an average
$7\%$ accuracy cost.


Combined \sysname{} (Dynamic$\times$Lossy) reaches $2.42\times$ throughput on average, the best of
all four configurations. Llama-3.1-8B illustrates both effects most clearly: each axis contributes
substantially in isolation ($1.71\times \to 2.23\times$ from dynamic draft tree shaping,
$1.71\times \to 2.29\times$ from lossy verification), and combining them yields $2.47\times$
($+44\%$) at $96\%$ recovery. The
Dynamic$\times$Lossy combination shows robustness, as this configuration delivers the best
speedup across every target model.


\begin{table}[t]
\caption{Ablation across draft-tree construction (static, dynamic) and verification rule
(strict, lossy), per target model and averaged across GSM8K, MATH-500, and HumanEval. Speedup is over vanilla AR; $\tau$ is the mean accepted token length; Rec.\,\% is accuracy
retained vs.\ EAGLE-3 $(3,1,4)$. All lossy rows use the single global $\kappa{=}0.2$. Best
speedup per row is in bold.}
\label{tab:2x2}
\centering
\small
\begin{tabularx}{\linewidth}{@{}l l *{6}{>{\centering\arraybackslash}X}@{}}
\toprule
& & \multicolumn{3}{c}{\textbf{Static tree}} & \multicolumn{3}{c}{\textbf{Dynamic tree}} \\
\cmidrule(lr){3-5}\cmidrule(lr){6-8}
\textbf{Model} & \textbf{Verif.} & Speedup & $\tau$ & Rec.\% & Speedup & $\tau$ & Rec.\% \\
\midrule
\multirow{2}{*}{Llama-3.1-8B}
 & Strict & 1.71 & 2.49 & 100 & 2.23 & 3.69 & 100 \\
 & Lossy  & 2.29 & 4.19 & 91  & \textbf{2.47} & 4.21 & 96 \\
\midrule
\multirow{2}{*}{Qwen3-8B}
 & Strict & 1.58 & 2.23 & 100 & 1.76 & 2.74 & 100 \\
 & Lossy  & 1.48 & 2.62 & 91  & \textbf{1.84} & 2.93 & 95 \\
\midrule
\multirow{2}{*}{DeepSeek-R1-8B}
 & Strict & 2.16 & 3.05 & 100 & 2.65 & 4.40 & 100 \\
 & Lossy  & 2.53 & 4.54 & 96  & \textbf{2.97} & 5.09 & 94 \\
\midrule
\multirow{2}{*}{Average}
 & Strict & 1.82 & 2.59 & 100 & 2.21 & 3.61 & 100 \\
 & Lossy  & 2.10 & 3.78 & 93  & \textbf{2.42} & 4.08 & 95 \\
\bottomrule
\end{tabularx}
\end{table}

\paragraph{Hyperparameter search.}
We evaluate each component of \sysname{} by searching its hyperparameters in isolation: the
divisor $d$ in \dcs{} on dynamic-only (lossless verification), and the margin threshold $\kappa$
on lossy-only (static chain $(7,1,8)$), followed by a final $\kappa$ search on the joint dynamic
and lossy configuration that represents \sysname{} (Combined). The divisor $d$ sets how quickly the confidence signal saturates the tree-shape range, $d{=}0.5$, gives the best throughput in every (model, benchmark)
cell; we fix $d{=}0.5$ throughout. The value $\kappa$ trades throughput against accuracy recovery
monotonically, and the trade-off has the same shape on all three targets, which is what makes a
single global setting viable: we adopt $\kappa{=}0.2$ for every cell of Table~\ref{tab:main}, keeping recovery without giving up significant throughput gain. The full per-cell sweeps for both hyperparameters are in
Appendix~\ref{sec:hyperparamsearch}.

\paragraph{Batch size.}
Table~\ref{tab:bs} reports throughput at batch sizes $1$-$16$ on DeepSeek-R1-8B across all three
benchmarks, with speedup measured against vanilla autoregressive decoding at the same batch size.
\sysname{} leads EAGLE-3 in every cell, and the absolute speculative speedup falls as the batch
grows for both methods, as the workload shifts from memory bandwidth-bound to
compute-bound~\citep{leviathan2023speculative}. The corresponding grids for Llama-3.1-8B and
Qwen3-8B are in Appendix~\ref{sec:bsfull}.

\begin{table}[H]
\centering
\caption{Throughput speedup over vanilla autoregressive decoding at the same batch size, on
DeepSeek-R1-8B across all three benchmarks, at the single global $\kappa{=}0.2$. \sysname{}
leads at every batch size on every benchmark.}
\label{tab:bs}
\small
\begin{tabular*}{\linewidth}{@{\extracolsep{\fill}}ll ccccc@{}}
\toprule
\textbf{Benchmark} & \textbf{Method} & \textbf{1} & \textbf{2} & \textbf{4} & \textbf{8} & \textbf{16} \\
\midrule
\multirow{2}{*}{GSM8K}
 & EAGLE-3 & 2.28 & 2.15 & 2.05 & 1.70 & 1.33 \\
 & \sysname{} (Combined) & \textbf{3.20} & \textbf{3.02} & \textbf{2.61} & \textbf{2.14} & \textbf{1.67} \\
\midrule
\multirow{2}{*}{MATH-500}
 & EAGLE-3 & 2.15 & 2.07 & 2.03 & 1.69 & 1.35 \\
 & \sysname{} (Combined) & \textbf{2.97} & \textbf{2.78} & \textbf{2.40} & \textbf{2.02} & \textbf{1.62} \\
\midrule
\multirow{2}{*}{HumanEval}
 & EAGLE-3 & 2.05 & 1.96 & 1.88 & 1.57 & 1.28 \\
 & \sysname{} (Combined) & \textbf{2.74} & \textbf{2.52} & \textbf{2.20} & \textbf{1.81} & \textbf{1.46} \\
\bottomrule
\end{tabular*}
\end{table}

\vspace{-2em}
\paragraph{Temperature $T{=}1$.}
We ran the evaluation on
Llama-3.1-8B at temperature $T{=}1$ over three seeds, \sysname{} (Combined) leads lossless
EAGLE-3 on all three benchmarks at the same temperature, by $+16\%$ on GSM8K, $+41\%$ on
MATH-500 and $+17\%$ on HumanEval, and raises the mean accepted length in every cell. Accuracy
recovery stays between $93\%$ and $98\%$, comparable to the greedy regime. The full grid,
including per-seed standard deviations, is in Appendix~\ref{sec:temperature}.

\paragraph{Experiment run variance.}
We run every (model, benchmark) cell of Table~\ref{tab:main} three times at the global
$\kappa{=}0.2$ (Appendix~\ref{sec:variance}). Under greedy decoding the generated sequence is
deterministic, so task accuracy, recovery and $\tau$ reproduce exactly across independent
launches. The only source of variation is end-to-end throughput, whose spread is up to $1.8\%$ across the cells.

\subsection{Limitation analysis of FLy's verifier}
\label{sec:windowfree}

FLy's window rule assumes that the lookahead window $w$ fits comfortably inside the typical
accepted prefix. This premise holds for the long draft chains ($\nsteps{=}15$-$25$) FLy was
designed for, but breaks on shallower drafters such as EAGLE-3. Measuring mean accepted token
length $\tau$ for static EAGLE-3 $(3,1,4)$ across falls strictly below FLy's $w{=}6$. Per target, $\tau$ averages $2.49$ on
Llama-3.1-8B, $2.23$ on Qwen3-8B and $3.05$ on DeepSeek-R1-8B. The full analysis of mean $\tau$ per configuration is in Appendix~\ref{sec:eagle3-mat}. This structural mismatch underlines the importance of \sysname's window-free design: by reading the target's distribution at the mismatch position rather than looking ahead, the verifier becomes independent of draft depth and applicable regardless of $\tau$.

\section{Conclusion}
\label{sec:conclusion}

We presented \sysname, a training-free per-step adaptive speculative decoder. Its primary contribution is a novel lossy verification rule that converts near-miss draft tokens into accepted tokens using only the target distribution at the verification step, without requiring an external verifier for equivalence checking. Another contribution is a per-step dynamic tree-shape policy that adjusts all three EAGLE-3 hyperparameters independently, removing the fixed total-budget assumption common to prior adaptive-tree work. The two components are empirically orthogonal axes of throughput and their gains compound: \sysname{} improves end-to-end throughput by 16-45\% on average per target model (peak +56\%), while retaining 94--96\% of downstream task accuracy on average per target model, all inside SGLang. Beyond raw efficiency, \sysname{} lowers the latency and energy cost of serving LLMs, broadening access to LLMs. 

\section{Future work}

\paragraph{Speculative Decoding Beyond Language Domain}
Any autoregressive architecture is constrained by the same sequential dependency, and the draft-then-verify pattern could be adapted beyond language tokens. A growing body of work applies Transformer-based architectures to embodied settings, such as vision-language-action models for robotics 
\cite{brohan2023rt2,kim2024openvla}. There, latency is not a matter of user experience but a requirement for effective coordination; an agent acting in the world cannot wait seconds between perceiving and moving. Although \sysname is specific to language, we believe its central idea, adapting speculative decoding to the per-step difficulty of the task, is a valuable foundation for extending the draft-then-verify paradigm to these physical domains.

\paragraph{Retrieval-Based Drafting for Agentic Workloads}
In agentic settings agents emit long traces that are repetitive across similar requests. This repetition allows sequences generated once to be reused in future steps. A line of work exploits this by replacing the draft model with retrieval, drawing candidate tokens from prior outputs, or a datastore rather than an autoregressive generation \cite{saxena2023promptlookup,he2024rest,luo2024tokenrecycling}. SuffixDecoding \cite{oliaro2024suffixdecoding} applies this to agentic workloads specifically, caching past sequences in a suffix tree.
These methods change the source of the draft tokens, whereas our \sysname adapts the draft tree shape and the verification rule. A direction for future work is replacing autoregressive drafters with retrieval-based drafters, and maintaining the \sysname's per-step signals adaptive mechanism to improve inference efficiency in repetitive agentic settings.

\section{Limitations}
\label{sec:limitations}
The per-step margin rule does not preserve the target distribution in the formal sense of \cite{leviathan2023speculative}; accuracy retention is reported empirically and may vary on tasks outside our evaluation. We build and evaluate \sysname on EAGLE-3 drafter and do not report results on diffusion-based drafters such as DFlash, DDTree~\citep{chen2026dflashblockdiffusionflash, ringel2026ddtree}, a recent SOTA diffusion tree drafter that appeared close to our submission. We believe that \sysname's lossy verification and the tree-shape principle would transfer to DDTree; empirical confirmation is left to future work.


\clearpage
\newpage

{
\small
\bibliographystyle{plainnat}
\bibliography{custom,anthology-1,anthology-2}
}








\appendix
\section{EAGLE3 target--draft model pairings}
\label{sec:appendixpairs}

\begin{table}[htbp]
\caption{Target--draft model pairings used in our experiments. The target model is listed on the first line of each pair and the EAGLE-3 draft checkpoint on the second.}
\centering
\footnotesize
\setlength{\tabcolsep}{0pt}
\begin{tabular}{@{}p{\columnwidth}@{}}
\toprule
\textbf{Target} \;/\; \textbf{Draft checkpoint} \\
\midrule
Llama \\
\texttt{meta-llama/Llama-3.1-8B-Instruct} \\
\texttt{lmsys/sglang-EAGLE3-LLaMA3.1-Instruct-8B} \\
\midrule
\addlinespace[3pt]
DeepSeek \\
\texttt{deepseek-ai/DeepSeek-R1-Distill-Llama-8B} \\
\texttt{yuhuili/EAGLE3-DeepSeek-R1-Distill-LLaMA-8B} \\
\midrule
\addlinespace[3pt]
Qwen \\
\texttt{Qwen/Qwen3-8B} \\
\texttt{AngelSlim/Qwen3-8B\_eagle3} \\
\bottomrule
\end{tabular}
\label{tab:model-pairings}
\end{table}

\section{Experiment configurations}

Each (method, model, benchmark) cell in Table~\ref{tab:main} completes in 2--6 hours on a single A100 (80\,GB), varying with benchmark output length.

\label{sec:experimentconfigs}
\begin{table}[htbp]
\caption[Method hyperparameter configurations used in Table \ref{tab:main}.]{Method hyperparameter configurations used in Table \ref{tab:main}. ``Tree shape'' is the triplet $(\nsteps, \topk, \ndt)$ or the dynamic range $A{\leftrightarrow}B$. $d$ is the DCS divisor; $\kappa$ margin threshold of the lossy margin rule. TALON: $\mu$ confidence-margin threshold. FLy: $w$ lookahead window, $\theta$ entropy threshold. The margin threshold is fixed globally at $\kappa{=}0.2$; the full sweep is reported in Appendix~\ref{sec:hyperparamsearch}. $^{*}$TALON and FLy are re-implementations on SGLang using the EAGLE-3 drafter.}
\centering
\footnotesize
\setlength{\tabcolsep}{0pt}
\begin{tabular}{@{}p{\columnwidth}@{}}
\toprule
\textbf{Method} \;/\; \textbf{Tree shape $(\nsteps, \topk, \ndt)$; hyperparameter details} \\
\midrule
EAGLE-3 \\
$(3,1,4)$; Strict \\
\midrule
\addlinespace[3pt]
TALON* \\
$(\nsteps,\topk)$ adaptable, $\ndt{=}8$ fixed; Strict; $\mu{=}0.03$ \\
\midrule
\addlinespace[3pt]
FLy* \\
$(7,1,8)$; Lossy (window); $\nsteps{=}7$, $w{=}3$, $\theta{=}0.30$ \\
\midrule
\addlinespace[3pt]
\sysname{} (Dyn-only) \\
$(3,4,4){\leftrightarrow}(7,1,8)$; Strict; $d{=}0.5$ \\
\midrule
\addlinespace[3pt]
\sysname{} (Lossy-only) \\
$(7,1,8)$; Lossy (margin);  $\kappa{=}0.2$ \\
\midrule
\addlinespace[3pt]
\sysname{} (Combined) \\
$(3,4,4){\leftrightarrow}(7,1,8)$; Lossy (margin); $d{=}0.5$, $\kappa{=}0.2$  \\
\bottomrule
\end{tabular}
\label{tab:method-configs}
\end{table}






\section{Hyperparameter search}
\label{sec:hyperparamsearch}
\paragraph{Draft tree shape ($d$).} The divisor $d$ controls how rapidly \dcs{} saturates the
tree-shape range. A small $d$ pushes the policy toward the deep-narrow extreme, producing chains
that maximise accept length when the draft is strong; a large $d$ keeps the policy in the
shallow-wide regime when the draft is uncertain. Table~\ref{tab:divisor-sweep} sweeps
$d \in \{0.2, 0.3, 0.4, 0.5\}$ on the Dynamic-only configuration. Throughput does not decrease
with $d$ in any of the nine cells, and $d{=}0.5$ is the best value in all of them, so we fix
$d{=}0.5$ for the Combined search and for every \sysname{} run in Table~\ref{tab:main}.

\begin{table}[htbp]
\caption[Divisor hyperparameter search for \sysname{}'s (Dynamic-only) draft tree construction over $d \in \{0.2, 0.3, 0.4, 0.5\}$.]{Divisor
hyperparameter search for \sysname{}'s (Dynamic-only) draft tree construction over
$d \in \{0.2, 0.3, 0.4, 0.5\}$. S is the speedup over vanilla autoregressive decoding and $\tau$
the mean accepted token length. Smaller $d$
saturates the policy toward the deep-narrow end of the $(3,4,4){\leftrightarrow}(7,1,8)$ range;
larger $d$ keeps it shallow-wide. As $d{=}0.5$ wins on every cell, we fix $d{=}0.5$ for all
\sysname{} (Combined) runs in Table~\ref{tab:main}.}
\label{tab:divisor-sweep}
\centering
\small
\begin{tabularx}{\linewidth}{@{}l l *{6}{>{\centering\arraybackslash}X}@{}}
\toprule
& & \multicolumn{2}{c}{GSM8K} & \multicolumn{2}{c}{MATH-500} & \multicolumn{2}{c}{HumanEval} \\
\cmidrule(lr){3-4}\cmidrule(lr){5-6}\cmidrule(lr){7-8}
\textbf{Model} & $d$ & S & $\tau$ & S & $\tau$ & S & $\tau$ \\
\midrule
\multirow{4}{*}{Llama-3.1-8B}
 & 0.2 & 2.02 & 3.68 & 1.93 & 3.21 & 2.34 & 3.86 \\
 & 0.3 & 2.15 & 3.74 & 1.99 & 3.23 & 2.36 & 3.91 \\
 & 0.4 & 2.18 & 3.77 & 2.03 & 3.28 & 2.41 & 3.96 \\
 & \textbf{0.5} & 2.22 & 3.80 & 2.05 & 3.28 & 2.43 & 3.99 \\
\midrule
\multirow{4}{*}{Qwen3-8B}
 & 0.2 & 1.76 & 2.90 & 1.52 & 2.44 & 1.56 & 2.50 \\
 & 0.3 & 1.84 & 2.97 & 1.59 & 2.50 & 1.65 & 2.61 \\
 & 0.4 & 1.89 & 3.02 & 1.61 & 2.52 & 1.69 & 2.64 \\
 & \textbf{0.5} & 1.91 & 3.03 & 1.64 & 2.54 & 1.72 & 2.65 \\
\midrule
\multirow{4}{*}{DeepSeek-R1-8B}
 & 0.2 & 2.84 & 4.95 & 2.56 & 4.30 & 2.27 & 3.81 \\
 & 0.3 & 2.85 & 4.97 & 2.63 & 4.38 & 2.33 & 3.85 \\
 & 0.4 & 2.89 & 4.99 & 2.63 & 4.35 & 2.36 & 3.88 \\
 & \textbf{0.5} & 2.90 & 4.98 & 2.64 & 4.32 & 2.41 & 3.90 \\
\bottomrule
\end{tabularx}
\end{table}

\paragraph{Lossy threshold ($\kappa$).} The threshold $\kappa$ controls how strictly the lossy
verifier promotes a draft token.
Table~\ref{tab:kappa-lossy-combined}(a) search over $\kappa \in \{0.10, 0.20, 0.30\}$ on the static
lossy chain $(7,1,8)$, and (b) repeats the search on the Combined configuration with
$d{=}0.5$ fixed. Both sections show the same throughput-recovery trade-off: lower $\kappa$
increases throughput and decreases recovery, higher $\kappa$ the reverse. The ordering is stable
across models and across both configurations, and $\kappa{=}0.2$ is the point at which recovery stays high while retaining most of the available throughput gain. 

\begin{table}[H]
\caption[Margin rule threshold hyperparameter search for \sysname{} over $\kappa \in \{0.05, 0.10, 0.20, 0.30\}$.]{Margin
rule threshold hyperparameter search for \sysname{} over $\kappa \in \{0.05, 0.10, 0.20, 0.30\}$,
reported for transparency; the adopted global value $\kappa{=}0.2$ (bold) is fixed once and used
unchanged in every cell of Table~\ref{tab:main}. S is the speedup over vanilla autoregressive
decoding, $\tau$ the mean accepted token length, R the accuracy retained relative to lossless
EAGLE-3. (a) Lossy-only with the fixed chain $(7,1,8)$; (b) Combined with $d{=}0.5$ and the
dynamic range $(3,4,4){\leftrightarrow}(7,1,8)$.}
\label{tab:kappa-lossy-combined}
\centering
\footnotesize
\renewcommand{\arraystretch}{0.95}
\begin{tabularx}{\linewidth}{@{}l l *{9}{>{\centering\arraybackslash}X}@{}}
\toprule
& & \multicolumn{3}{c}{\textbf{GSM8K}} & \multicolumn{3}{c}{\textbf{MATH-500}} & \multicolumn{3}{c}{\textbf{HumanEval}} \\
\cmidrule(lr){3-5}\cmidrule(lr){6-8}\cmidrule(lr){9-11}
\textbf{Model} & $\kappa$ & S & $\tau$ & R & S & $\tau$ & R & S & $\tau$ & R \\
\midrule
\multicolumn{11}{@{}l}{\textbf{(a) \sysname{} (Lossy-only)}: search over $\kappa$, chain $(7,1,8)$}\\
\midrule
\multirow{3}{*}{Llama-3.1-8B}
 & 0.10 & 2.48 & 4.66 & 88 & 1.91 & 3.31 & 80  & 2.70 & 4.86 & 89 \\
 & \textbf{0.20} & 2.28 & 4.35 & 93 & 1.85 & 3.33 & 86  & 2.75 & 4.89 & 94 \\
 & 0.30 & 2.22 & 4.17 & 95 & 1.63 & 3.28 & 92  & 2.68 & 4.79 & 98 \\
\midrule
\multirow{3}{*}{Qwen3-8B}
 & 0.10 & 1.66 & 2.95 & 98 & 1.13 & 2.51 & 76  & 1.56 & 2.82 & 84 \\
 & \textbf{0.20} & 1.61 & 2.85 & 98 & 1.34 & 2.38 & 82  & 1.49 & 2.64 & 93 \\
 & 0.30 & 1.57 & 2.78 & 99 & 1.32 & 2.33 & 79  & 1.40 & 2.50 & 96 \\
\midrule
\multirow{3}{*}{DeepSeek-R1-8B}
 & 0.10 & 3.34 & 6.08 & 93 & 2.93 & 5.24 & 97  & 2.77 & 4.53 & 89 \\
 & \textbf{0.20} & 3.20 & 5.78 & 96 & 1.80 & 3.28 & 111 & 2.58 & 4.55 & 82 \\
 & 0.30 & 3.14 & 5.60 & 99 & 1.66 & 3.20 & 109 & 2.48 & 4.42 & 82 \\
\midrule
\multicolumn{11}{@{}l}{\textbf{(b) \sysname{} (Combined)}: $d{=}0.5$, search over $\kappa$}\\
\midrule
\multirow{4}{*}{Llama-3.1-8B}
 & 0.05 & 2.64 & 4.72 & 89  & 2.26 & 3.72 & 71  & 2.77 & 4.70 & 94 \\
 & 0.10 & 2.55 & 4.55 & 91  & 2.19 & 3.59 & 84  & 2.57 & 4.35 & 91 \\
 & \textbf{0.20} & 2.43 & 4.32 & 94  & 2.15 & 3.51 & 90  & 2.82 & 4.80 & 103 \\
 & 0.30 & 2.40 & 4.22 & 94  & 2.13 & 3.47 & 96  & 2.78 & 4.74 & 102 \\
\midrule
\multirow{4}{*}{Qwen3-8B}
 & 0.05 & 2.03 & 3.29 & 100 & 1.80 & 2.86 & 79  & 1.93 & 3.08 & 89 \\
 & 0.10 & 2.00 & 3.23 & 99  & 1.78 & 2.82 & 79  & 1.89 & 3.02 & 98 \\
 & \textbf{0.20} & 1.95 & 3.17 & 99  & 1.74 & 2.72 & 82  & 1.82 & 2.90 & 104 \\
 & 0.30 & 1.94 & 3.12 & 98  & 1.68 & 2.66 & 83  & 1.77 & 2.82 & 109 \\
\midrule
\multirow{4}{*}{DeepSeek-R1-8B}
 & 0.05 & 3.43 & 6.14 & 89  & 3.09 & 5.24 & 91  & 2.93 & 4.93 & 82 \\
 & 0.10 & 3.35 & 5.93 & 92  & 2.18 & 3.56 & 111 & 2.83 & 4.72 & 83 \\
 & \textbf{0.20} & 3.20 & 5.66 & 97  & 2.97 & 5.01 & 98  & 2.74 & 4.60 & 88 \\
 & 0.30 & 3.14 & 5.51 & 97  & 2.11 & 3.53 & 105 & 2.64 & 4.39 & 92 \\
\bottomrule
\end{tabularx}
\end{table}

\section{Batch-size evaluation}
\label{sec:bsfull}
Analysis of \sysname{} performance across varying batch-size. All runs use the single global $\kappa{=}0.2$;
speedup is over vanilla autoregressive decoding at the corresponding batch size.

\begin{table}[H]
\centering
\caption{Throughput speedup over vanilla autoregressive decoding at the same batch size, on
Llama-3.1-8B, at the single global $\kappa{=}0.2$. Best per cell in bold. \sysname{} leads
throughout on GSM8K and HumanEval; on MATH-500, where mean accepted length is lowest, EAGLE-3
is ahead from batch size $4$ upward.}
\label{tab:bs-llama}
\small
\begin{tabular*}{\linewidth}{@{\extracolsep{\fill}}ll ccccc@{}}
\toprule
\textbf{Benchmark} & \textbf{Method} & \textbf{1} & \textbf{2} & \textbf{4} & \textbf{8} & \textbf{16} \\
\midrule
\multirow{2}{*}{GSM8K}
 & EAGLE-3 & 1.93 & 1.84 & 1.72 & 1.42 & 1.14 \\
 & \sysname{} (Combined) & \textbf{2.43} & \textbf{2.31} & \textbf{1.90} & \textbf{1.61} & \textbf{1.26} \\
\midrule
\multirow{2}{*}{MATH-500}
 & EAGLE-3 & 1.38 & 1.77 & \textbf{1.70} & \textbf{1.46} & \textbf{1.20} \\
 & \sysname{} (Combined) & \textbf{2.15} & \textbf{1.99} & 1.68 & 1.38 & 1.13 \\
\midrule
\multirow{2}{*}{HumanEval}
 & EAGLE-3 & 1.81 & 1.93 & 1.60 & 1.38 & 1.04 \\
 & \sysname{} (Combined) & \textbf{2.82} & \textbf{2.63} & \textbf{2.35} & \textbf{2.33} & \textbf{2.66} \\
\bottomrule
\end{tabular*}
\end{table}

\begin{table}[H]
\centering
\caption{Throughput speedup over vanilla autoregressive decoding at the same batch size, on
Qwen3-8B, at the single global $\kappa{=}0.2$. Best per cell in bold. \sysname{} leads up to
batch size $8$ on all three benchmarks; at batch size $16$ the two converge and EAGLE-3 is
marginally ahead on GSM8K and MATH-500.}
\label{tab:bs-qwen}
\small
\begin{tabular*}{\linewidth}{@{\extracolsep{\fill}}ll ccccc@{}}
\toprule
\textbf{Benchmark} & \textbf{Method} & \textbf{1} & \textbf{2} & \textbf{4} & \textbf{8} & \textbf{16} \\
\midrule
\multirow{2}{*}{GSM8K}
 & EAGLE-3 & 1.68 & 1.65 & 1.59 & 1.36 & \textbf{1.16} \\
 & \sysname{} (Combined) & \textbf{1.95} & \textbf{1.86} & \textbf{1.62} & \textbf{1.40} & 1.08 \\
\midrule
\multirow{2}{*}{MATH-500}
 & EAGLE-3 & 1.56 & 1.45 & 1.41 & 1.24 & \textbf{1.06} \\
 & \sysname{} (Combined) & \textbf{1.74} & \textbf{1.67} & \textbf{1.48} & \textbf{1.31} & 1.03 \\
\midrule
\multirow{2}{*}{HumanEval}
 & EAGLE-3 & 1.50 & 1.49 & 1.46 & 1.28 & 1.09 \\
 & \sysname{} (Combined) & \textbf{1.82} & \textbf{1.75} & \textbf{1.59} & \textbf{1.41} & \textbf{1.10} \\
\bottomrule
\end{tabular*}
\end{table}

\section{Temperature $T{=}1$ analysis}
\label{sec:temperature}
Table~\ref{tab:temperature} shows the evaluation at temperature
$T{=}1$ on Llama-3.1-8B over three seeds; speedup is over autoregressive decoding and recovery
is relative to lossless EAGLE-3, both at the same temperature.

\begin{table}[H]
\centering
\caption{\sysname{} (Combined) under temperature $T{=}1$ on Llama-3.1-8B, at the single global
$\kappa{=}0.2$, over three seeds (mean $\pm$ std). Speedup is over autoregressive decoding and
recovery is relative to lossless EAGLE-3, both at the same temperature. Under $T{=}1$ the
generated sequence is no longer deterministic, so $\tau$ and recovery also carry a standard
deviation.}
\label{tab:temperature}
\small
\begin{tabular*}{\linewidth}{@{\extracolsep{\fill}}ll ccc@{}}
\toprule
\textbf{Benchmark} & \textbf{Method} & \textbf{Speedup} & \textbf{MAT\,$\tau$} & \textbf{Rec.\%} \\
\midrule
\multirow{2}{*}{GSM8K}
 & EAGLE-3 (lossless) & $1.65 \pm 0.01$ & $2.35 \pm 0.01$ & 100 (ref) \\
 & \sysname{} (Combined) & $\mathbf{1.91 \pm 0.07}$ & $\mathbf{3.28 \pm 0.07}$ & $98.1 \pm 2.2$ \\
\midrule
\multirow{2}{*}{MATH-500}
 & EAGLE-3 (lossless) & $1.11 \pm 0.02$ & $1.55 \pm 0.02$ & 100 (ref) \\
 & \sysname{} (Combined) & $\mathbf{1.56 \pm 0.06}$ & $\mathbf{2.56 \pm 0.05}$ & $94.6 \pm 3.1$ \\
\midrule
\multirow{2}{*}{HumanEval}
 & EAGLE-3 (lossless) & $1.95 \pm 0.00$ & $2.76 \pm 0.01$ & 100 (ref) \\
 & \sysname{} (Combined) & $\mathbf{2.29 \pm 0.07}$ & $\mathbf{4.05 \pm 0.03}$ & $93.2 \pm 2.0$ \\
\bottomrule
\end{tabular*}
\end{table}

\section{Experiment run variance}
\label{sec:variance}
Table~\ref{tab:variance} shows the variance experiment, by repeating Table~\ref{tab:main} runs three times as independent
runs at the global $\kappa{=}0.2$. Under greedy decoding the generated sequence is
deterministic, so task accuracy, recovery and $\tau$ are identical across launches and are
reported as point values; only end-to-end throughput varies, and we report it as mean $\pm$
standard deviation. The spread is up to $1.8\%$
across cells. 
\begin{table}[htbp]
\centering
\caption{Run-to-run variance of \sysname{} (Combined) at the global $\kappa{=}0.2$, batch
size~1, greedy decoding, three independent runs per cell. Speedup over autoregressive decoding
is reported as mean $\pm$ std; $\tau$ and recovery are deterministic under greedy decoding and
are reported as point values.}
\label{tab:variance}
\small
\begin{tabular*}{\linewidth}{@{\extracolsep{\fill}}ll ccc@{}}
\toprule
\textbf{Model} & \textbf{Benchmark} & \textbf{Speedup} & \textbf{MAT\,$\tau$} & \textbf{Rec.\%} \\
\midrule
\multirow{3}{*}{Llama-3.1-8B}
 & GSM8K     & $2.43 \pm 0.01$ & 4.32 & 94 \\
 & MATH-500  & $2.15 \pm 0.01$ & 3.51 & 90 \\
 & HumanEval & $2.82 \pm 0.05$ & 4.80 & 103 \\
\midrule
\multirow{3}{*}{Qwen3-8B}
 & GSM8K     & $1.95 \pm 0.00$ & 3.17 & 99 \\
 & MATH-500  & $1.74 \pm 0.00$ & 2.72 & 82 \\
 & HumanEval & $1.82 \pm 0.01$ & 2.90 & 104 \\
\midrule
\multirow{3}{*}{DeepSeek-R1-8B}
 & GSM8K     & $3.20 \pm 0.03$ & 5.66 & 97 \\
 & MATH-500  & $2.97 \pm 0.00$ & 5.01 & 98 \\
 & HumanEval & $2.74 \pm 0.01$ & 4.60 & 88 \\
\bottomrule
\end{tabular*}
\end{table}

\section{Mean accepted token length for static EAGLE-3}
\label{sec:eagle3-mat}
Per-cell values behind the analysis in Section~\ref{sec:windowfree}. All runs use the static
EAGLE-3 configuration $(3,1,4)$ under strict verification.

\begin{table}[H]
\centering
\caption[Mean accepted token length $\tau$ for static EAGLE-3 across models and benchmarks,
showing the limitation of the FLy method.]{Mean accepted token length $\tau$ for static EAGLE-3
$(3,1,4)$ across the three target models and three benchmarks. $\tau$ ranges from $2.08$ to
$3.27$, in every case well short of the length $w{=}6$ that FLy's lookahead window requires.}
\label{tab:eagle3-mat}
\footnotesize
\begin{tabularx}{\linewidth}{@{}l *{3}{>{\centering\arraybackslash}X}@{}}
\toprule
\textbf{Target Model} & \textbf{GSM8K $\tau$} & \textbf{MATH-500 $\tau$} & \textbf{HumanEval $\tau$} \\
\midrule
Llama-3.1-8B    & 2.83 & 2.08 & 2.55 \\
Qwen3-8B        & 2.39 & 2.21 & 2.10 \\
DeepSeek-R1-8B  & 3.27 & 3.02 & 2.86 \\
\bottomrule
\end{tabularx}
\end{table}





\end{document}